\documentclass{sn-jnl}

\usepackage{amsmath,amsthm,amssymb,amsfonts}
\usepackage{graphicx}
\usepackage[colorinlistoftodos]{todonotes}

\newcommand{\R}{\mathbb{R}}
\newcommand{\N}{\mathbb{N}}

\usepackage[font=small,labelfont=bf]{caption}
\usepackage{enumerate}
\usepackage{dirtytalk}

\usepackage[autostyle]{csquotes}
\usepackage{hyperref}
\usepackage{ulem}
\usepackage{csquotes}
\usepackage{ dsfont }
\usepackage{subcaption}

\begin{document}

\title[Towards Generalized Entropic Sparsification \\ for Convolutional Neural Networks]{Towards Generalized Entropic Sparsification \\ for Convolutional Neural Networks}


\author*[1]{\fnm{Tin} \sur{Barisin}}\email{lastname@rptu.de}

\author[1]{\fnm{Illia} \sur{Horenko}}\email{lastname@rptu.dem}

\affil[1]{\orgdiv{Mathematics Department}, \orgname{RPTU Kaiserslautern-Landau}, \orgaddress{\street{Gottlieb-Daimler-Straße 47}, \city{Kaiserslautern}, \postcode{67663}, 
\country{Germany}}}


\abstract{Convolutional neural networks (CNNs) are reported to be overparametrized. The search for optimal (minimal) and sufficient architecture is an NP-hard problem as the hyperparameter space for possible network configurations is vast. 
Here, we introduce a layer-by-layer data-driven pruning method based on the mathematical idea aiming at a computationally-scalable entropic relaxation of the pruning problem. The sparse subnetwork is found from the pre-trained (full) CNN using the network entropy minimization as a sparsity constraint. This allows deploying a numerically scalable algorithm with a sublinear scaling cost. The method is validated on several benchmarks (architectures): (i) MNIST (LeNet) with sparsity $55\%-84\%$ and loss in accuracy $0.1\%-0.5\%$, and (ii) CIFAR-10 (VGG-16, ResNet18) with sparsity $73-89\%$ and loss in accuracy $0.1\%-0.5\%$. 
}




\maketitle

\section{Introduction}
Deep learning has enabled significant progress in processing signals, images, speech, text, and other data modalities by capturing and detecting complex patterns in the data. 
Mathematical validity of deep networks is dwelling on the universal approximation theorems \cite{siegel2020approximation,JCM-38-502}, where a problem of learning is cast as a problem of function approximation, when a composition of stacked parameterized linear and non-linear operators is selected to minimize the pre-defined loss function - measuring the function approximation quality - on a given data using gradient-based optimization methods.
On a practical side, these advances have been achieved using large datasets, very deep models with large number of parameters, large computing centers with many GPUs, and huge involvement of hardware and energy resources in the training phase. For example, ChatGPT foundation model GPT-3 \cite{brown2020language} is reported to have 175 billion parameters and required $3.14\times 10^{23}$ floating point operations for training with $300$ billion tokens on V100 GPU high-bandwidth cluster. There are some evidences that the further developments of large lange models require several order of magnitude more hardware and energy resources then the last published evidences about the GPT-3 model. 


In the next phase, it is expected that these models will be scaled and optimized so that they can bring benefits to a wide range of applications.
One thing that always follows deep neural networks is the large number of parameters, which is tied to the high memory requirements and high energy consumption. Hence, optimizing for these factors would bring additional value to the field, especially in use cases where the models are deployed on large scales across many memory-constrained devices. 
One way to tackle this problem is to reduce the model complexity, i.e. to find a model with a smaller number of parameters and floating point operations that mimic the performance of their larger counterpart. This can be seen as a problem of finding the optimal architecture of the deep network for a given problem. 
Systematic grid search across all possible architectural designs of the deep networks is an NP-hard problem, and is prohibitively expensive in time and computational resources.  
Hence, the researchers resort to the technique known as network pruning (or sparsification): take a large pre-trained model and remove the model components that have minimal effect on the model performance.
The outcome is often the significantly smaller model in terms of parameters with comparable performance. 

\begin{figure}
    \centering
    \begin{subfigure}{0.54\textwidth}
    \includegraphics[width=\textwidth]{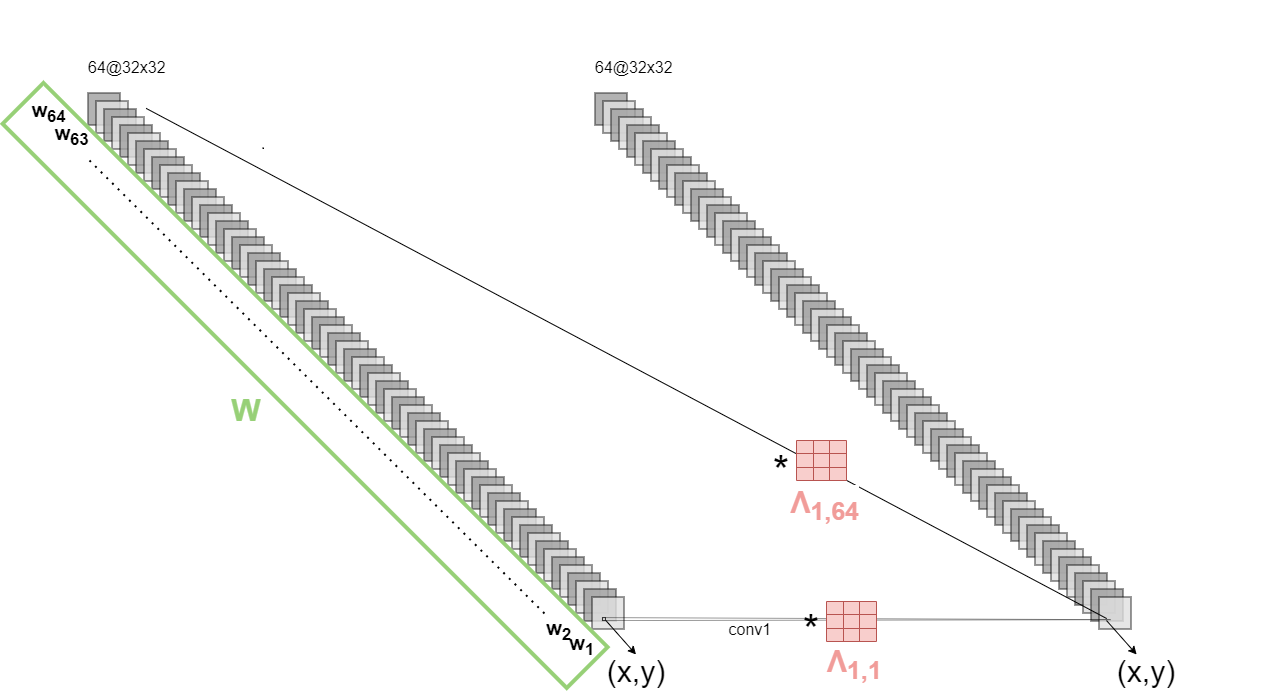}
    \caption{Convolutional layer conv1 from VGG-16 (Table \ref{tab:VGG-overview}).}
    \end{subfigure}
    \centering
    \begin{subfigure}{0.44\textwidth}
    \includegraphics[width=\textwidth]{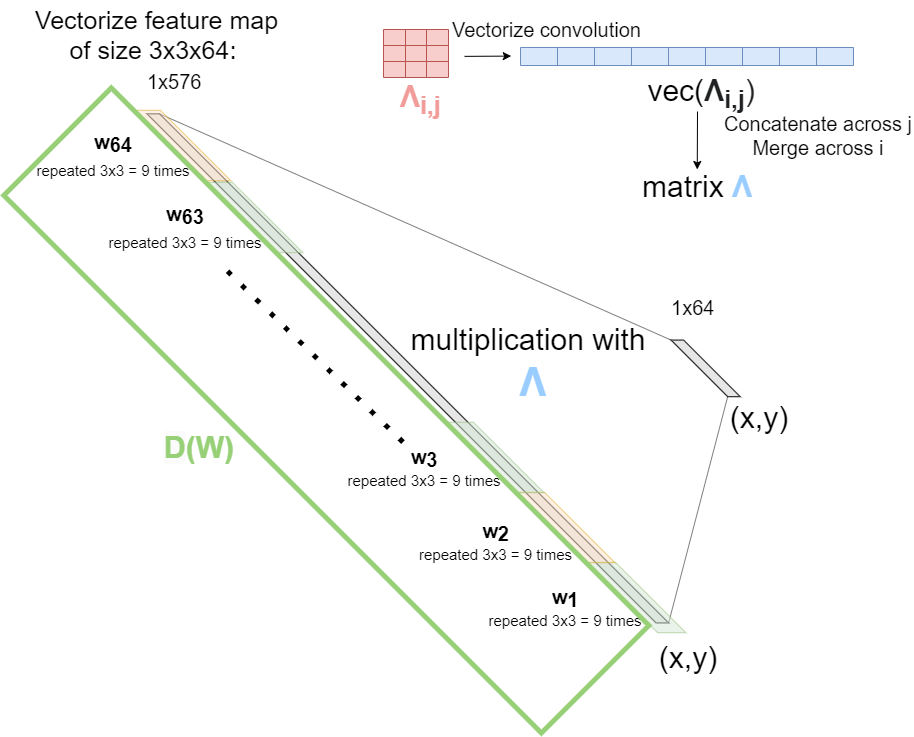}
    \caption{Interpreting conv1 as a linear layer.}
    \end{subfigure}
    \caption{Illustration: transforming convolutional layer conv1 from VGG-16 (Table \ref{tab:VGG-overview}, Appendix \ref{ref:net:details}) as a fully connected layer in every point $(x,y)$ from the image domain, and applying sparsification method by sparsely solving the linear system of equation in $D(W)$ and $\Lambda$ based on SPARTAn algorithm, see (\ref{spartan_eq}) from Section \ref{sec:gesCONV}. Layer conv1 transforms the feature map of dimension $64\times32\times32$ to the output of the same dimension using $64\times64$ convolutions with masks of size $3\times 3$. }
    \label{fig:intro}
\end{figure}

We contribute to this area by introducing a data-driven layer-by-layer sparsification method based on the linear adaptation of the non-linear regression method known as SPARTAn \cite{horenko2023cheap2}.
The main idea behind the proposed approach is to use entropic regularization \cite{horenko2023cheap2,horenko2020scalable,vecchi2022espa+} on the input channels of each layer while using them as regressors to predict the corresponding output channels. Here, minimizing entropy can be interpreted as a sparsity constraint in the regression problem.
This is a direct generalization of the previous work on the sparsification of the fully connected deep networks \cite{barisin2024entropy1}.
Contributions of this work are summarized in the following points.
 \begin{itemize}
     \item The SPARTAn  algorithm \cite{horenko2023cheap2} is adapted to sparsify the convolutional layers (Figure \ref{fig:intro}). 
     We show that all the benefits of the SPARTAn algorithm, such as sublinear cost scaling and an ability to work on the "small data", directly transfer to convolutional layers with arbitrary support, too. 
    \item We validate the goodness of the proposed approach on MNIST and CIFAR-10 datasets and various convolutional networks such as LeNet, VGG-16, and ResNet. For example, $89\%$ of weights from VGG-16 can removed with minimal loss in the performance ($<0.1\%$) on CIFAR-10. Furthermore, we determine which layers have the most redundancies and can be pruned more than the others.
    \item We investigate the claim that the value of network pruning is in the fact that it can discover optimal network architecture, while the given weights are of the less of the value. The reason is that training from scratch from randomly initialized weights would give a comparable or even better performance. We verify that training from scratch gives equivalent results. However, it requires more training epochs than with fine tuning. Hence, this proves the usefulness of keeping the weights from the pre-trained sparsified model.

 \end{itemize}

\subsection{Entropy and machine learning}
In information theory, the concept of entropy was introduced by Claude Shannon \cite{shannon1948mathematical} to measure the amount of information or uncertainty that can be encoded in the signal during the communication between the source and the receiver. 
In the words of the probability theory, entropy is defined as an expected self-information of the random variable. 
Formally, let $w=(w_1,...,w_d) \in \R^d_{\geq 0}$, $\sum_{i=1}^d w_i = 1$ be a discrete probability vector, discrete Shannon entropy is defined 
\begin{equation}
    \label{eq:shannon}
    H(w) = - \sum_{i=1}^d w_i \log w_i.
\end{equation} 
The uniform distribution is the distribution with the highest uncertainty, i.e. with the maximal entropy. On the other hand, entropy minimal distribution is the deterministic one, i.e. it is a distribution with a single possible outcome.

Entropy has found its way into the field of machine learning, as well. 
For example, maximizing entropy across feature dimensions encourages more features to be used in the algorithm rather than less, which reduces the risk of overfitting \cite{meister2020, horenko2020scalable}. On the other hand, entropy minimization favors the reduction of the feature space dimension preferring simpler models \cite{barisin2024entropy1}. 
As neural networks are argued to be overparameterized \cite{denil2013predicting}, we focus on entropy minimization with the aim of finding smaller models without the loss in performance.
Alternatively, entropy-based techniques have also been applied to the problem of robust adversarial training \cite{jagatap2022adversarially} or for semi-supervised learning \cite{grandvalet2004semi}.

\textbf{Entropic learning} is a new class of machine learning methods designed to tackle the problems of overfitting and lack of robust classifier. They are proven to be effective even in small data problems where the dimension of feature space is larger than the size of the training set. 
The central idea behind entropic learning is to use the entropy function (\ref{eq:shannon}) as a regularization on the probability vector $w$ that is introduced on the feature space. This term can be flexibly added in the loss function and is controlled by hyperparameter $\epsilon_w$. 
Entropy-optimal scalable probabilistic approximation (eSPA) \cite{horenko2020scalable}, its extension eSPA+ \cite{vecchi2022espa+}, and Sparse Probabilistic Approximation for Regression Task Analysis (SPARTAn) \cite{horenko2023cheap2} represent this class of methods. 
eSPA \cite{horenko2020scalable} and eSPA+ \cite{vecchi2022espa+} are designed for classification tasks and optimize the loss function in four variables $(w,S,\Gamma,\Lambda)$. The loss function consists of entropic regularization, Kullback-Leibler divergence to measure classification error, and the box-discretization error term. It is minimized using a four-step algorithm\footnote{Each step of algorithm optimizes in one of the variables $\{w,S,\Gamma,\Lambda\}$, while keeping the remaining ones fixed.} which is proven to have a low computational cost and to be monotonous convergent (Theorem 1 in \cite{horenko2020scalable}). The difference between the two methods is in the fact that eSPA+ \cite{vecchi2022espa+} has a closed-form analytical solution for every step of the optimization algorithm, in contrast to eSPA which uses a convex optimization solver to optimize for $w$. SPARTAn \cite{horenko2023cheap2} is the adaptation of the same algorithm but for the regression task.

\textbf{Entropy based methods for network pruning:}
As entropy measures the amount of information, it can be deployed to measure the amount of information on the neuron level in deep networks and use it to distinguish more informative neurons from the less informative ones. Hence, as this naturally leads to network pruning or sparsification, we briefly revise the work based on this idea, e.g. \cite{luo2017entropy, hur2019entropy, li2019using, li2019exploiting, wang2021filter}.
Luo et al \cite{luo2017entropy} calculate the entropy on the output probability distribution for every filter. The low entropy of the filter implies a low information amount and it is pruned.
Li et al \cite{li2019using} and Wang et al \cite{wang2021filter} derive the heuristical filter importance score based on the layer output entropy for every image and filter separately. 
Alternatively, Hur et al \cite{hur2019entropy} use the entropy on the weight distribution
to derive the adaptive threshold for iterative pruning of the individual weights with too low entropy.
Liu et al \cite{li2019exploiting} introduce a kernel sparsity and entropy (KSE) indicator, a data-driven entropy indicator calculated on the $k$-nearest neighbour distance distribution of the filter values.  

Our method differs from the existing work, as the entropy is not used to quantify the amount of information of the filter, but rather employs an entropic regularization in the interpretation of the layer-by-layer sparsification problem as a regression learning task. The minimum entropy principle enforces the sparsity of the solution and prunes the less contributing filters.

\subsection{Related work: network prunning}
In this section, we revise the pruning methods from the literature and illustrate the principles behind the pruning heuristics.
Generally, the methods are divided based on the aim of the pruning algorithm. For example, individual weights can be pruned (\textit{unstructured pruning}), the whole neurons or convolutional filters (\textit{structured pruning}). Alternatively, 
inference of the layer can be done faster when the
the weight matrix is decomposed in lower dimensional components which enable faster computation (\textit{low rank decomposition}).

Unstructured pruning methods require custom implementation for sparse operations to achieve faster inference. Individual weights can be pruned on pre-trained models by thresholding \cite{han2015learning} or during training using sparsity constraint in the loss function \cite{retsinas2020, liu2015}.
For structured pruning methods it is easier to achieve the benefit of faster inference time without any adaptations, as its result is a smaller network in terms of the number of filters (neurons) per layer. 
It is often possible to design both structured and unstructured pruning methods using a certain pruning principle or technique. However, structured pruning methods remain preferred as it is easier to achieve the main benefits of the pruning such as faster inference time or lower memory consumption.
For example, $L_1$ regularization is an efficient way to control the norm of the weights (or filters) in the linear regression tasks, e.g. Lasso regression. Hence, imposing $L_1$ regularization on the network's weights has been a popular technique to prune the network in various settings and adaptations \cite{li2016pruning,he2017,wen2016,kumar2021pruning,liu2017learning, guo2023automatic}.
Alternatively, network pruning can be interpreted as a dimensionality reduction problem, and hence, Principal Component Analysis (PCA) \cite{garg2019low} or Eucledian projection \cite{ma2019resnet} can be used to design the pruning criteria. 
Furthermore, a variety of other structure pruning techniques exists, e.g. variational Bayesian pruning \cite{zhao2019variational}, pruning with reinforcement learning \cite{wei2022automatic}, layer-wise relevance propagation \cite{aketi2020gradual}, information bottleneck \cite{guo2023automatic}, etc.
Low rank decomposition methods apply the various results for matrix decomposition from linear algebra, e.g. singular value decomposition (SVD) \cite{denton2014}, generalized singular value decomposition (GSVD) \cite{zhang2015}, approximation using the product of two lower dimensional matrices \cite{denil2013predicting}, approximation using a linear combination of separable 1d basis filters \cite{jaderberg2014}.
Interestingly, Denil et al \cite{denil2013predicting} make a hypothesis that the weight matrix can be predicted just from the few elements of the same matrix and use their method to empirically prove the claim.

Some works in the literature re-evaluate the value of network pruning, as an alternative to the variety of successful pruning techniques. 
Liu et al \cite{liu2018rethinking} state that pruning can be seen as a method for finding an optimal network architecture, as smaller models trained from scratch obtain comparable or even better performance than the pruned models. 
This would imply that using the random search of the architecture one can discover smaller models. This idea was explored in \cite{li2022revisiting} by randomly selecting the number of channels in each layer. This way many random subnetworks are evaluated in a greedy fashion and the best-performing ones are chosen for fine tuning.

\section{Method}
In this section we introduce the linear adaptation of the SPARTAn algorithm for convolutional layers.
Sparse Probabilistic Approximation for Regression Task Analysis (SPARTAn) \cite{horenko2023cheap2} is a method for solving a non-linear regression problem. 
It is based on the partition of the feature space with $K$ discretization boxes and approximation with $K$ piecewise linear regression problems with entropic regularization on the feature space.
More details on SPARTAn are given in Appendix \ref{spartan_recap}.
The linear adaptation of SPARTAn is given by setting $K=1$ and solving a single linear regression task with entropic regularization (Appendix \ref{sec:fcc:layer:sparse}). 
By minimizing entropy this regression task is constrained to result in a sparse solution. 
\\
Prior to that, we show how the convolutional layer can be interpreted as a linear (fully connected) layer (Section \ref{sec:problemForm}) and formulate the problem of the sparsification task for the convolutional layer of the deep network (Section \ref{sec:problem:formulation}). This formulation requires the adaptation in the SPARTAn algorithm. The probabilistic vector $w$ on the input (feature) dimension needs to be applied on the whole convolutional filters rather than on individual weights, i.e. to perform structured pruning rather than unstructured. This is described in detail in Section \ref{sec:gesCONV}.


\subsection{Notation and basics on convolutional layers in deep networks }
\label{sec:problemForm}

Convolutional layers are the basic building blocks of convolutional neural networks (CNNs) and the main source of a large number of parameters, as they are stacked one upon another and combined with non-linearity to achieve a highly parameterized non-linear classifier. In this section we revise the concepts behind the convolutional layers.

Images are discrete representations of continuous functions with the compact support from $\mathcal{L}_2(\R^2)$, i.e. 2d image can be seen as $I: \{1,\cdots, H\} \times \{1, \cdots, W \} \to \R$, or in our notation as $H \times W$ matrices $I \in \R^{H\times W}$ for $H,W \in \N$. 
In discrete setting, convolution of image $I \in \R^{H\times W}$ with kernel $Q\in \R^{k\times k}$, with $k=2l+1 \in \N$ is defined 
\begin{equation}
\label{eq:discrete:conv}
    (I \ast Q) (x_1,x_2) = \sum_{y_1=-(k-1)/2}^{(k-1)/2} \sum_{y_2=-(k-1)/2}^{(k-1)/2} I(x_1-y_1,x_2-y_2) Q(y_1,y_2). 
\end{equation}

Generally, the convolutional layer takes a stack of 2d feature maps as an input, applies convolution operation (\ref{eq:discrete:conv}) to every feature map, and finally sums them to calculate the 2d output feature map. As output is again multidimensional, this is repeated for every 2d output feature map.
Kernels $Q$ from (\ref{eq:discrete:conv}) are made of trainable parameters and are defined on compact support for $k=3,5$. These parameters are learned through training procedure by optimizing a loss function. In the remainder of this section, we formally introduce the convolutional layer as this is needed to formulate the sparsification problem in Section \ref{sec:problem:formulation}.

For that purpose let $D\in \N$ be an input (feature) dimension and let $M\in \N$ be an output dimension of the layer. This means that the layer transforms a $D$-dimensional input feature tensor into a $M$-dimensional output tensor. 
Let $T\in \N$ be a number of data points. Furthermore, let $X_{t}=(X_{1,t}, \cdots, X_{D,t} ) \in \big(\R^{H\times W}\big)^{D}$ be an instance of the input data, i.e. explanatory variables and let
$Y_{t}=(Y_{1,t}, \cdots, Y_{M,t} ) \in \big(\R^{H\times W}\big)^{M}$ 
be a corresponding instance of the output data, i.e. response variables for $t \in \{1,...,T \}$.  
Then, a convolutional layer\footnote{Convolutional layers also have a learnable bias parameter which we omit for the sake of the brevity.} of kernel size $k$ 
can be seen as a convolution with $Q =(Q_{i,j})_{i,j=1}^{D,M} \in \big(\R^{k\times k}\big)^{D\times M}$ 

\begin{equation}
    \label{convolutional:layer}
    Y_{j,t} = \sum_{i=1}^D X_{i,t} \ast Q_{i,j}
\end{equation} 
for $j = 1,...,M$. Next, we show that the convolutional layer can be written as a linear layer of fully connected network. This would enable us to apply the entropic sparsification algorithm for linear layers.

\subsection{Problem formulation: sparsification for convolutional layers}
\label{sec:problem:formulation}

Similarly as in \cite{barisin2024entropy1}, we formulate the sparsification problem as the removal task of the input dimensions which contribute the least to calculating the output feature maps.
The idea is that by removing the input dimensions, we also remove the corresponding kernels from $Q$ and consequently reduce the number of parameters in the layer.
Hence, the sparsification objective becomes finding $\hat{D} << D$, subset of indices $S \subset \{1,\cdots, D\}$ with $|S| = \hat{D}$, and convolutional kernels $\hat{Q} =(\hat{Q}_{i,j})_{i,j=1}^{\hat{D},M} \in \big(\R^{k\times k}\big)^{\hat{D}\times M}$
that approximate $Y_{j,t}$ from (\ref{convolutional:layer}) by
\begin{equation}
    \label{convolutional:layer:sparse}
    \hat{Y}_{j,t} = \sum_{i\in S} X_{i,t} \ast \hat{Q}_{i,j}
\end{equation} 
for $j = 1,...,M.$ That is, the goal is to balance between the sparsity of representation $\hat{D}$ and the approximation error $\sum_{t=1}^{T}||\hat{Y}_{t} - Y_{t}||$. 
Next, we show how this approximation problem for the convolutional layer can be interpreted as a linear regression task.

\subsubsection{Interpreting convolutional layer as linear layer}
\label{sec:conv:layer:simplification}

The convolutional layer can be interpreted as a linear layer in each point $(x_1, x_2)$ from the image domain, i.e. it can be shown that the following holds
\begin{equation}
\label{conv:layer:simplification}
     Y_{t}(x_1,x_2) = Q^{vec} \text{ }A(X_{t})(x_1,x_2).
\end{equation}
Here, $Q^{vec} = \Big(Q^{vec}_{1}, \cdots, Q^{vec}_{M} \Big) \in \R^{M\times (k^2 D)}$ is a kernel weight matrix where $Q^{vec}_{j} \in \R^{k^2D}$ is a vectorized kernel $Q_{j} \in (\R^{k\times k})^D$ for $j=1,\cdots,M$.
Vector $A(X_{t})(x_1,x_2)\in \R^{k^2 D}$ is a vector of feature values from $X_t$ for $(x_1,x_2)$ and their neighbours with distance $k$ in $||\cdot||_\infty$ norm.
A detailed derivation of this result is given in Appendix \ref{conv:layer:linear}.

In the context of convolutional layers, every point from the image domain $H\times W$ becomes one data point to solve the system of linear equations, i.e. we can say that the actual number of data points is $T^* = (H\times W) \times T$. 
Since the convolutional layer can be interpreted as a linear layer, this problem (\ref{convolutional:layer:sparse}) can be seen as a sparse entropic regression task.

\subsection{Generalized entropic sparsification for convolutional layers}
\label{sec:gesCONV}

Although we have seen in Section \ref{sec:conv:layer:simplification} that the convolutional layer can be interpreted as a linear layer, that does not mean that we can directly apply the linear adaptation of SPARTAn (Appendix \ref{sec:fcc:layer:sparse}) to this problem. The reason is that the sparsification objective is set on the convolutional filter level, see (\ref{convolutional:layer:sparse}). Hence, we introduce the sparse entropic regression for the convolutional layers, a generalized linear adaptation of the SPARTAn algorithm for convolutional layers.

In the next sections we will use the simplified notation which can be achieved through the variable substitution in (\ref{conv:layer:simplification}): $ (t,x_1,x_2) \to t$, $A(X_{t})(x_1,x_2) \to X_t$, and $T^* \to T$, i.e. the following sparse entropic regression task will be solved for the following data  
\begin{align*}
    &\{(X_t, Y_t) \text{ } | \text{ } t = 1, \cdots, T\}
    \\
    & \text{with } X_t \in \R^{k^2D},  Y_t \in \R^M.
\end{align*}

\subsubsection{Sparse entropic regression for convolutional layers}
Sparse entropic regression for convolutional layers generalizes the linear adaptation of the SPARTAn algorithm
using a discrete probability vector $w=(w_1,...,w_D) \in \R^D_{\geq 0}$, $\sum_{i=1}^D w_i = 1$ and a matrix $\Lambda \in \R^{M\times(k^2D)}$ so that the following loss function is minimized
\begin{align}
\label{spartan_eq}
\begin{split}
\mathcal{L}_{sparsify}(w, \Lambda) &=  \epsilon_w\underbrace{\sum_{d=1}^D w_d \log w_d}_{\text{Entropy}} + 
\epsilon_{l_2} \underbrace{\sum_{m,d=1}^{M,(k^2D+1)} \Lambda^2_{m,d-1}}_{L_2}+ \\
&+ \underbrace{\frac{1}{TM} \sum_{t,m=1}^{T,M} \Bigg( Y_{m,t} - \Lambda_{m,0} - \sum_{d=1}^D w_d \bigg(\sum_{l=1}^{k^2} \Lambda_{m,(d-1)k^2+l}X_{(d-1)k^2+l,t} \bigg) \Bigg)^2}_{MSE},
\end{split}
\end{align}
with respect to chosen hyperparameters $(\epsilon_w, \epsilon_{l_2})$. 
Loss function $\mathcal{L}_{sparsify}$ consists of three terms: mean square (or approximation) error or (MSE), $L_2$ regularization on matrix $\Lambda$, and entropic regularization on probability vector $w$. The sparsity of the vector $w$ is ensured from the minimum entropy principle in entropy regularization, i.e. $\epsilon_w < 0$. 

Minimization of $\mathcal{L}_{sparsify}$ is achieved using a two-step algorithm. The principle is as follows: each step of the algorithm keeps one variable ($\Lambda$ or $w$) fixed, and minimizes the loss in the remaining one ($w$ and $\Lambda$, respectively).
This gives a monotonous decrease in the loss function. These two steps are alternated until the convergence is achieved, i.e. until the difference in loss function for two iterations of the algorithm is smaller than the predefined tolerance level.
Details on derivations of $w$-step and $\Lambda$-step of optimization algorithm are described in Appendix~\ref{optimize_sparsification}.

Vector $w$ has a probabilistic interpretation as a feature selection (or importance) vector on $X_t$.
Essentially, the idea behind the generalization is that convolutional filters belonging to the same input channel $i \in \{1,\cdots,D\}$ are grouped and assigned a joint weight $w_i$. Alternatively, this can be interpreted again as sparsification of the feature space of dimension $D$ that are represented as input channels (Section~\ref{sec:problem:formulation}).
\\
The regression weight matrix is given by
\begin{equation}
\label{eq:Qhat}
    \hat{Q}^{vec} =  \Lambda D(w)
\end{equation}
where 
$$D(w) = diag(\underbrace{w_1,...,w_1}_{k^2 \text{ times}}, \cdots, \underbrace{w_D,...,w_D}_{k^2 \text{ times}} ) \in  \R^{ (k^2D) \times (k^2D)}.$$
Feature dimensions, whose vector components $w_i$ are close to zero (e.g. $<10^{-6}$), have no effect on the approximation error (MSE) and hence can be discarded. 
Hence, in this way we are able to reduce the number\footnote{Note that $\hat{Q}^{vec}$ from (\ref{eq:Qhat}) is still $(M \times k^2D)$ matrix rather than $(M \times k^2\hat{D})$ for $\hat{D} << D$. However, it is sparse, i.e. it has $(D-\hat{D})$ null columns which are removed together with the respective rows from $X_t$ as they do not contribute to calculating $\hat{Y}_t$. After removing the null columns from $\hat{Q}^{vec}$, one can transform it back to the form of kernels, and calculate it as in (\ref{convolutional:layer:sparse}).} of the input channels to $\hat{D} << D$. 


\subsubsection{Connection to the entropic sparsification of the fully connected layers}
Fully connected layers linearly transform the features only across the feature dimension, in contrast to the convolutional layers that additionally use the spatial dimensions of the fixed window size $k\times k$ (e.g. $3\times 3$ or $5\times 5$). For this reason fully connected layers are also sometimes referred to as convolutional layers on the window of size $1\times 1$ ($k=1$), or 1d convolutional layers.
Hence, within the introduced framework and notation, sparsification of the fully connected layers becomes a special case of the sparsification of the convolutional layers for $k=1$ and represents a direct generalization of the previous work \cite{barisin2024entropy1}.

\section{Experiments}
We use the following metric to measure the sparsity of the network:
$$ \text{Sparsity(model)} 
\frac{\# \text{parameters(baseline model)} - \# \text{parameters(model)}}{\# \text{parameters(baseline model)}}.$$
It represents the percentage of weights of the baseline model that were removed through sparsification. This measure can also be calculated on every layer (or subparts of the network) separately giving us insight into the local sparsity of the network.

Another measure that we will use in Section \ref{sec:sparsifyVGG} is FLOPs (Floating Point Operations). FLOPs capture the number of arithmetic operations required to perform an inference step for one image. It serves as a measure of computation complexity and energy consumption.

\subsection{Sparsifying LeNet on MNIST}
\begin{figure}
    \centering
    \includegraphics[width = \textwidth]{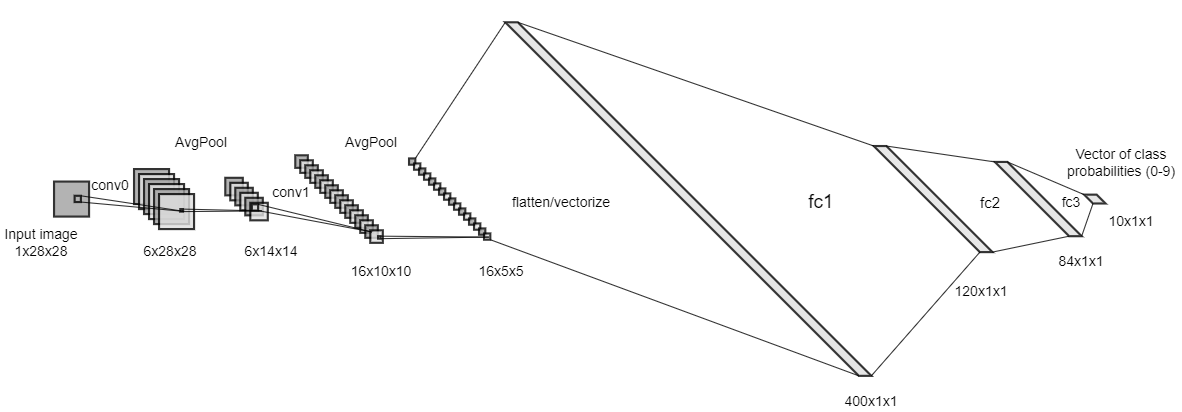}
    \caption{LeNet \cite{lecun98} is an example of a convolutional neural network that consists of both convolutional and fully connected layers. This figure visualizes an architecture described in Table \ref{tab:lenet} (Appendix \ref{ref:net:details}). Both convolutional layers are defined on the window of size $5\times5$.}
    \label{fig:lenet:architecture}
\end{figure}

LeNet \cite{lecun98} is one of the first convolutional neural networks that has surpassed 99\% accuracy on the MNIST data set. It consists of two convolutional and three fully connected layers. The convolutional layer uses a mask of size $5\times 5$. Overall, the network has in total 61,706 parameters. The network architecture is described in Table \ref{tab:lenet} (Appendix \ref{ref:net:details}) or visualized in Figure \ref{fig:lenet:architecture}.

MNIST data set \cite{lecun98} consists of images of digits between 0 and 9 belonging to one of ten classes. The dataset has in total $60,000$ images of size $28\times 28$ for training and validation, and additional  $10,000$ images for testing. We used $50,000$ images for training, and $10,000$ images for validation.

The network was trained for 20 epochs using ADAM optimizer \cite{kingma2014adam} with a starting learning rate of $0.001$ which is cut in half every 7 epochs.
After sparsification, the fine tuning was performed for $10$ epochs with a learning rate of $0.0001$ which is cut in half every 4 epochs.

\begin{table}[]
    \centering
    \begin{tabular}{|c|c|c|c|c|}
        \hline
         Model &  conv1 \#output channels & accuracy (fine tuned) & \#parameters & sparsity\\
         \hline
         baseline & 16 & 99.04\% (-) & 61,706 & 0\% \\
         E1 & 8 & 98.49\% (99.03\%) & 36,498 & 40.85\% \\
         E2 & 4 & 97.35\% (98.73\%) & 23,894 & 61.28\% \\
         \hline
    \end{tabular}
    \caption{Sparsifying only the convolutional layer of LeNet. Accuracy after fine tuning is reported in the brackets of the accuracy column. Configuration E1 denotes $(\epsilon_w, \epsilon_{l_2}) =(-0.01,0.01)$ and E2 denotes $(\epsilon_w, \epsilon_{l_2}) =(-0.05,0.05)$.}
    \label{tab:lenet:exp1}
\end{table}

\begin{table}[]
    \centering
    \begin{tabular}{|c|c|c|c|c|}
        \hline
         Model &  fc1-fc3 \#output channels & accuracy (fine tuned) & \#parameters & sparsity \\
         \hline
         baseline & 400-120-84 & 99.04\% (-) & 61,706 & 0\% \\
         E1+fc-v1 & 200-104-43 & 97.87\% (98.92\%) & 27,223 & 55.89\% \\
         E1+fc-v2 &  200-40-18 & 96.89\% (98.47\%) & 10,332 & 83.25\% \\
         \hline
    \end{tabular}
    \caption{Sparsifying the whole LeNet: Accuracy after fine tuning is reported in the brackets of the accuracy column. Configuration E1 denotes $(\epsilon_w, \epsilon_{l_2}) =(-0.01,0.01)$. For fully-connected layers we used the following configurations: $(\epsilon_w,\epsilon_{l_2}) =(-0.0001,0.0001)$ for v1 and $(\epsilon_w,\epsilon_{l_2}) =(-0.001,0.001)$ for v2.}
    \label{tab:lenet:exp:full}
\end{table}

\textbf{Sparsifying only convolutional layer:} 
We sparsify only the output of conv1 from Table \ref{tab:lenet} (Appendix \ref{ref:net:details}) to validate the goodness of our method. This layer has $16$ output channels.
LeNet uses $5\times5$ convolutions, hence $k=5$ and $X_t \in \R^{25\times 16}$ with 16-dimensional feature probability vector $w=(w_1,...,w_{16})$. The results are given in Table \ref{tab:lenet:exp1}. By cutting the number of channels in half (E1$ \to (\epsilon_w, \epsilon_{l_2}) =(-0.01,0.01)$) and quarter (E2$ \to (\epsilon_w, \epsilon_{l_2}) =(-0.05,0.05)$), we are able to achieve identical or almost identical accuracy (difference of $0.3\%)$ after fine tuning while removing roughly $40\%$ and $60\%$ of network parameters, respectively.

\textbf{Sparsifying the whole network:} 
Next, we aim to sparsify the remaining two fully connected layers of the network. Here, we use the version of this algorithm for $k=1$, which is a generalized version of the entropic sparsification from \cite{barisin2024entropy1}. For fully connected layers we use the following two configurations of hyperparameters
$(\epsilon_w,\epsilon_{l_2}) =(-0.0001,0.0001)$ for v1 and $(\epsilon_w,\epsilon_{l_2}) =(-0.001,0.001)$ for v2, and combine them with E1 from earlier to get the fully sparsified model.
The results are given in Table \ref{tab:lenet:exp:full}.
Finally, one can remove between $55\%$ and $84\%$ of parameters, with a decrease of performance between $0.1\%$ and $0.5\%$, respectively. 

\subsection{CIFAR-10}
CIFAR-10 \cite{krizhevsky2010cifar} is a dataset based on photographs of objects and animals with varying positions and backgrounds. This dataset consists of 60,000 images of size $32\times 32$ divided into 10 classes: airplane, automobile, bird, cat, deer, dog, frog, horse, ship, and truck. 
CIFAR-10 is split into 6 batches with 10,000 images, out of which one batch is reserved for testing. 

\subsubsection{Sparsifying VGG-16} 
\label{sec:sparsifyVGG}

\begin{table}[]
\hspace{-1cm}
    \centering
    \begin{tabular}{|c|c|}
        \hline
        Model & Architecture \\
        \hline
        baseline & $64, 64, \text{M}, 128, 128, \text{M}, 256, 256, 256, \text{M}, 512, 512, 512, \text{M}, 512, 512, 512, \text{M}$  
        \\
        \hline
        conv1-4 & $29, 64, \text{M}, 124, 127, \text{M}, 256, 256, 256, \text{M}, 512, 512, 512, \text{M}, 512, 512, 512, \text{M}$
        \\ 
        conv5-7 & $64, 64, \text{M}, 128, 128, \text{M}, 250, 232, 219, \text{M}, 512, 512, 512, \text{M}, 512, 512, 512, \text{M}$ 
        \\
        conv8-10 & $64, 64, \text{M}, 128, 128, \text{M}, 256, 256, 256, \text{M}, 65, 24, 12, \text{M}, 512, 512, 512, \text{M}$ 
        \\
        conv11-12 & $64, 64, \text{M}, 128, 128, \text{M}, 256, 256, 256,\text{M}, 512, 512, 512, \text{M}, 10, 12, 512, \text{M}$ 
        \\
        \hline 
        conv8-12 & $64, 64, \text{M}, 128, 128, \text{M}, 256, 256, 256,\text{M}, 65, 24, 12, \text{M}, 10, 12, 512, \text{M}$ 
        \\
        conv\_all & $29, 64, \text{M}, 124, 127, \text{M}, 250, 232, 219, \text{M}, 65, 24, 12, \text{M}, 10, 12, 512, \text{M}$ 
        \\
        \hline
        conv\_all+fc &  $29, 64, \text{M}, 124, 127, \text{M}, 250, 232, 219, \text{M}, 65, 24, 12, \text{M}, 10, 12, 91, \text{M}$  \\
        \hline
    \end{tabular}
    \caption{Sparsified architectures of VGG-16 networks when selectively sparsifying either all or group of convolutional neural networks for the same parameter configuration $ (\epsilon_w, \epsilon_{l_2}) = (-0.0001, 0.0001)$. 
    }
    \label{tab_experiment_setting}
\end{table}
 
VGG-16 \cite{simonyan2015very} is a very deep neural network which consists of 13 convolutional and 3 fully connected layers. This network achieved state-of-the-art results on ImageNet in 2014. We used a slightly modified (and often used) version of VGG-16 with only 1 fully connected layer in order to reduce the number of parameters. Nevertheless, even with the modification the number of parameters is still large: 14.7 million. All convolutional layers are defined on $3\times 3$ mask. 
The overview of the architecture can be found in Table \ref{tab:VGG-overview} (Appendix \ref{ref:net:details}).

The network was trained for 200 epochs using Stochastic Gradient Descent (SGD) with a learning rate $0.1$, momentum set to $0.9$, and weight decay of $5\times10^{-4}$. The learning rate is adjusted using cosine annealing. We used the open-source implementation of VGG-16 in PyTorch\footnote{\url{https://github.com/kuangliu/pytorch-cifar/tree/master}}. The baseline model achieves $94.08\%$ accuracy on the test set.
Fine tuning was done using a learning rate $0.001$ for either $10$ or $50$ epochs, depending on the sparsification scenario from Table \ref{tab_experiment_setting}. When we sparsify a smaller subset of layers, we noticed that $10$ epochs were enough. For the fully sparsified network we used $50$ epochs.

For sparsification of every layer we used the same configuration of hyperparameters $(\epsilon_w, \epsilon_{l_2}) = (-0.0001,0.0001)$. Note that conv0 layer is never sparsified as it has only 3 input channels (corresponding to RGB images) and has an overall low number of parameters: less than 2,000. Each layer is sparsified from the baseline model independently of each other with $T=500$ data points.

\begin{table}[]
    \centering
    \begin{tabular}{|c|c|c|c|c|c|c|}
        \hline
        Model & Accuracy & \#parameters & Sparsity & Local sparsity &  Speedup & FLOPs\\ 
        & (fine tuned) & & & & (inference time) & (per image) \\
        \hline
        baseline & 94.08 \% & $1.47 \times 10^7$ & - & -  & - & $3.14 \times 10^8$ 
        \\
        \hline
        conv1-4 & 93.60\% & $1.47 \times 10^7$ & 0.02\% & 10.9\%  & 1.04x  & $2.91 \times 10^8$ 
        \\ 
        conv5-7 & 90.87\% (93.64\%) & $1.47 \times 10^7$ & 2.5\% & 16.99\% &  1.04x  & $2.98 \times 10^8$ 
        \\ 
        conv8-10 & 92.64\% (93.6\%) & $6.7\times 10^6$ & 54.59\% &  99.65\%  & 1.4x  & $2.14 \times 10^8$ 
        \\ 
        conv11-12 & 93.85\% & $7.7 \times 10^6$ & 47.38\% & 99.95\%  & 1.11x  & $2.87 \times 10^8$ 
        \\ 
        \hline 
        conv8-12 & 91.87\% (93.51\%) & $2.0 \times 10^6$ & 86.63\% & 99.50\% &  1.52x  & $1.95 \times 10^8$ 
        \\ 
        conv\_all & 83.52\% (93.45\%) & $1.7 \times 10^6$ & 88.4\% & - &  1.64x  & $1.57\times 10^8$ 
        \\
        \hline
        conv\_all+fc & 83.40\% (93.38\%) & $1.66 \times 10^6$ & 88.7\% & - &  1.67x  & $1.57 \times 10^8$ 
        \\
        \hline
    \end{tabular}
    \caption{Sparsified architectures of VGG-16 networks when selectively sparsifying either all or group of convolutional neural networks for the same parameter configuration $ (\epsilon_w, \epsilon_{l_2})= (-0.0001, 0.0001)$. Local sparsity indicates the redundancy of the layers based on the depth of the network. 
    In the second column fine tuned accuracy is reported in the brackets.}
    \label{tab_experiment_result}
\end{table}

\textbf{Sparsifying group of layers:}
First, we perform sparsification on the group of the layers. The goal is to analyze the local sparsity of the three groups of layers, i.e. to check whether there is more redundancy at certain parts of the network. It is important to note that the majority of parameters of the network are located towards the end (Table \ref{tab:VGG-overview} in Appendix \ref{ref:net:details}). This could indicate overparameterization of that part of the network.
Details on the architectures of the sparsified models are shown in 
Table \ref{tab_experiment_setting}. Table \ref{tab_experiment_result} shows that in the last two groups of layers almost $99.5\%$ of their weights can be removed, while for the first two only significantly fewer parameters seem to be redundant: between $10\%$ and $17\%$.

\textbf{Sparsifying the whole network:} 
The last three rows of Table \ref{tab_experiment_result} report the combined results when sparsifying the majority of convolutional layers (conv8-12), all convolutional layers (conv\_all), and all layers of the networks including fully connected ones (conv\_all+fc). In these scenarios one can achieve sparsity between $86\%$ and $88\%$ while preserving accuracy above $93.3\%$. Hence, the majority of weights can be removed with the performance drop of maximally $0.7\%$.
\\
A fully sparsified network cuts the runtime by $40\%$ and takes only 6.5MB in memory. The baseline model occupies significantly more memory: 57.5MB.
The total number of floating point operations (FLOPs) at inference step for fully sparsified network is cut in half compared to the (unsparsified) baseline (Table \ref{tab_experiment_result}). Hence, the sparisifed model significantly require the memory and computational requirements for model storage and inference.

\begin{table}[]
    \centering
    \begin{tabular}{|c||c|c|c|c|c|c|c|c|c|}
    \hline
     & epoch $0$ & epoch $1$ & epoch $10$ & epoch $20$ & epoch $50$ & epoch $70$ & epoch $100$ & epoch $160$  \\
    \hline
         fine tune & 83.40\% & 92.39\% & 93.07\% & 93.28\% & 93.38\% & 93.46\% & 93.66\% & 93.87\% \\
    \hline
         scratch & 10.0\% & 41.65\% & 76.64\% & 81.95\% & 83.86\% & 85.16\% & 85.56\% & 87.63\%  \\
    \hline
    \end{tabular}
    \caption{Sparsification of VGG-16. First row: effect of additional fine tuning on conv\_all+fc sparsified model from Table \ref{tab_experiment_result}. Second row: training the same network from scratch. Epoch $0$ refers to the non fine tuned network (first row) and randomly-initialized network (second row).}
    \label{tab:cifar10:epochs}
\end{table}
\begin{table}[]
    \centering
    \begin{tabular}{|c||c|c|c|}
        \hline
         &  epoch 200 & epoch 250 & epoch 280  \\
         \hline
         scratch & 90.88\% & 93.77\% & 94.02\% \\
         \hline
    \end{tabular}
    \caption{Training VGG-16 network from the scratch beyond $160$ epochs (Table \ref{tab:cifar10:epochs}). }
    \label{tab:cifar10:scratch}
\end{table}
\textbf{The effect on the number of epochs of fine tuning:} 
In previous experiments, we limited ourselves to fine tuning with $10$ and $50$ epochs. Here, we explore whether increasing the number of epochs can improve the results, as suggested by \cite{kumar2021pruning}, but with a learning rate which is 100 times smaller than the original training one.  Table \ref{tab:cifar10:epochs} shows that maximal accuracy is given after $160$ epochs, which indicates that accuracies of the models from Table \ref{tab_experiment_result} could be improved with additional fine tuning.
\\
The outcome of this experiment is also aligned with the hypothesis from \cite{liu2018rethinking}, which claims that the usefulness of the pruning methods lies in finding the optimal minimal architecture of the network which can be extensively fine tuned to recover the original accuracy. This is also aligned with criticism from \cite{li2022revisiting} which claims that the high performance of pruned network comes from prolonged fine tuning. 
However, our method reports only $1\%$ decrease in performance after 10 epochs of fine tuning, which is relatively modest fine tuning in comparison to 200 epochs that were used for training the full model. A single epoch of fine tuning results in $1.5\%$ decrease in performance.

\textbf{Training the sparse model from scratch:}
We further investigate the implications from \cite{liu2018rethinking} on whether the network pruning should be seen as a method for optimal network architecture search. 
Hence, a network with the same architecture as a fully sparisified one (last row of the Table \ref{tab_experiment_setting}) is trained from scratch using the same training setup as for training the original model. 
Table \ref{tab:cifar10:epochs} gives the comparison between fine tuned network and network trained from scratch for $160$ epochs. Table \ref{tab:cifar10:scratch} extends the experiment to $260$ epochs.
The outcome confirms the finding from \cite{liu2018rethinking}: a network trained from scratch can indeed reach the comparable accuracy to the fine tuned one. However, the cost of training from scratch is the increased number of epochs that is required: $100$ epochs more compared to the fine tuned network. 

\begin{table}[]
    \centering
    \begin{tabular}{|c|c|c|c|c|}
         \hline
         Method & Baseline & Sparsified model & \#parameters & Sparsity \\
        & accuracy & accuracy & (sparsified model) & 
         \\
         \hline
         Liu (L2) \cite{li2022revisiting} &94.63\% & 93.94\% & $6.2\times10^6$ & 57.89\% 
         \\
         Hao Li \cite{li2016pruning} & 93.25\% & 93.40\% & $5.40\times 10^6$ & 64.0\%  \\
         Liu (L1) \cite{li2022revisiting} & 94.63\% & 93.90\% & $5.05\times10^6$ & 65.68\%
         \\
         Zhao \cite{zhao2019variational} & 93.25\% & 93.18\% & $3.92\times 10^6$ & 73.34\%  \\
         Gard \cite{garg2019low} & 94.07\% & 93.76\% & $3.78\times 10^6$ & $74.2\%$ \\
         Guo \cite{guo2023automatic} & 93.68\% & 94.08\% & $\sim 3\times10^6$ & 78\%
         \\
         Liu \cite{liu2017learning} & 93.66\% & 93.80\% & $2.30\times 10^6$  & 88.5\% \\
         Aketi \cite{aketi2020gradual} & 93.95\% & 92.95\% & $1.36 \times 10^6$ & 91\% \\
         Kumar \cite{kumar2021pruning} & 93.77\% & 93.81\% & $1.07 \times 10^6$ & 92.7\%   \\
         Li \cite{li2019using} & 93.72\% & 93.76\% & $1.04 \times 10^6$ & 92.9\%  \\
         \hline
         ours & 94.08\% & 93.87\% & $1.66 \times 10^6$ & 88.7\% 
         \\
         ours (scratch) & 94.08\% & 94.02\% & $1.66 \times 10^6$ & 88.7\% 
         \\
         \hline
    \end{tabular}
    \caption{Comparison with other sparsification methods applied to VVG-16 on CIFAR-10. Methods are sorted by sparsity (last column). Note that Zhao \cite{zhao2019variational} uses 2 additional fully-connected layers on their version of VGG-16 model. The baseline model in Liu \cite{liu2017learning} has around $2\times 10^7$ parameters.}
    \label{tab:cifar10:comparison}
\end{table}

\textbf{Comparison with other methods:}
Table \ref{tab:cifar10:comparison} compares our methods with several methods from the literature. Our method achieves a slightly higher accuracy (by a small margin) than the remaining methods (except \cite{guo2023automatic}). Regarding sparsity only two methods \cite{kumar2021pruning, li2019using} achieve higher sparsity and comparable performance, but use fine tuning with a significantly higher learning rate. 
However, we also have not explored ideas such as finding the optimal hyperparameter configuration for $(\epsilon_w, \epsilon_{l_2})$, sparsifying more early layers as they have lower local sparsity (Table \ref{tab_experiment_result}). Here, we would benefit from numericaly-scalable optimization algorithm with a sublinear scaling of the cost.

\subsubsection{Sparsifying ResNet18} 

\begin{table}[]
\hspace{-1cm}
    \centering
    \begin{tabular}{|c|c|c|c|}
        \hline
        Model & Accuracy & \#parameters & Sparsity \\
        \hline
        baseline &  95.39\% & $ 1.1\times 10^7$ & -
        \\
        \hline
        conv(14,16)& 93.77\% & $ 3.9\times 10^6$ & 65.2\% \\
        (fine tuned: 1 epoch) & 94.7\% &  & 
        \\ %
        (fine tuned: 20 epochs) & 95.1\% & &  
        \\ %
        \hline
        conv\_all & 88.19\% & $ 2.9\times 10^6$ & 73.6\% 
        \\
        (fine tuned: 1 epoch) & 94.03\% &  &  
        \\
        (fine tuned: 50 epochs) & 94.91\% &  & 
        \\ %
        \hline
    \end{tabular}
    \caption{Sparsified architectures of ResNet18 networks when selectively sparsifying either all (\textit{conv\_all}) or only final group of layers ($\textit{conv(14,16)}$) with the same parameter configuration $ (\epsilon_w, \epsilon_{l_2})= (-0.0001, 0.0001)$. This indicates the redundancy of the layers based on the depth of the network. Scenario \textit{conv\_all} denotes sparsifying the following layers $(6,8,10,12,14,16)$ from Table \ref{tab:resnet18-overview} (Appendix \ref{ref:net:details}). In these  number of input channels from selected convolutional layers is reduced from $(128,128,256,256,512,512)$ to $(122,92,228,94,110,23)$.}
    \label{tab_experiment_result_resnet}
\end{table}

ResNet \cite{he2016deep} is a popular convolutional neural network that uses residual connections between convolutional blocks, see e.g. Figure \ref{fig:resblock} (Appendix \ref{ref:net:details}). 
This type of architecture tackles the accuracy degradation problem for deep networks when increasing their depth (number of layers).
We experiment with ResNet18 which consists of 18 $3\times 3$ convolutional layers and 3 shortcut $1\times 1$ layers. 
This network has 4 extra convolutional layers compared to VGG-16 and roughly 2 million parameters less. The baseline model is trained with the same setup as VGG-16 and achieves an accuracy of $95.37\%$ using 11,173,962 parameters.
An overview of the architecture of ResNet18 can be found in Table \ref{tab:resnet18-overview} (Appendix~\ref{ref:net:details}).

For entropic sparsification we used the same configuration as in the previous section: same hyperparameter configuration $ (\epsilon_w, \epsilon_{l_2})= (-0.0001, 0.0001)$. For simplicity, we focus on sparsifying only convolutional layers whose input is not directly used to build a residual connection (e.g. conv6 from Figure \ref{fig:resblock}). 
The reason for that is the following: the input in layers with (identity) residual connection contributes to two convolutional layers (e.g. conv5 and conv5shortcut from Figure \ref{fig:resblock} in Appendix \ref{ref:net:details}). This means by sparsifying two layers we are sparsifying this input twice. However, this could be resolved by taking the elementwise maximum for both $w$ vectors and re-estimating weights for both $\Lambda$s.

\begin{table}[]
    \centering
    \begin{tabular}{|c|c|c|c|}
        \hline
        Method & Baseline & Sparsified model & Sparsity \\
        & accuracy & accuracy &  \\
        \hline
        Wei \cite{wei2022automatic} & 94.87\% & 94.61\% & 30\%  \\
        Wei \cite{wei2022automatic} & 94.87\% & 94.87\% & 50\%  \\
        Wei \cite{wei2022automatic} & 94.87\% & 93.87\% & 70\%  \\
        Ma \cite{ma2019resnet} & 94.14\% & 93.22\% & 98.8\% \\
        \hline
         ours & 95.39\% & 94.91\% & 73.6\% \\
         \hline
    \end{tabular}
    \caption{Comparison with methods from the literature on ResNet18 network. Methods are sorted by sparsity (last column). Wei et al \cite{wei2022automatic} report results without fine tuning.}
    \label{tab:resnet18:comparison}
\end{table}

\textbf{Sparsification analysis and fine tuning:} 
We again perform two experiments: sparsifying only the last block of convolutional layers as they have the majority of parameters $(\tilde 70\%)$ and sparsifying all blocks except for the first one.
Table \ref{tab_experiment_result_resnet} reports the results of two sparsification scenarios. Sparsifying only the last block removes already $65\%$ of weights with a minimal drop ($0.3\%$) in performance after fine tuning with 20 epochs. When sparsifying the remaining blocks one can remove 1 million extra parameters with a slightly larger decrease ($0.5\%$ in the accuracy.
The memory storage requirements of baseline model are 42.6MB, while the (final) sparsified model requires significantly less memory: 11.3MB.

 \textbf{Comparison with methods from literature:}
Table \ref{tab:resnet18:comparison} shows comparison with methods from the literature. Note that we have not reported the results for the methods that do not give the number of parameters after sparsification, e.g. 
\cite{ning2020dsa, liu2020autocompress}. 
Our method achieves higher accuracies than the methods from Table \ref{tab:resnet18:comparison}. However, Wei et al \cite{wei2022automatic} do not apply fine tuning. On the other hand, Ma et al \cite{ma2019resnet} achieve significantly fewer parameters than our methods.
Note also that we only sparsify convolutional layers which do not belong to the residual connections. Hence, our method could potentially achieve higher sparsity rates as well.


\section{Discussion}
In this work, we introduced the generalized linear adaptation of SPARTAn algorithm. This algorithm is applied to data-driven layer-by-layer sparsification of convolutional layers from pre-trained CNNs. 
Sparsification is enabled by entropic regularization, i.e. by minimizing entropy on probability vector $w$ applied to the input feature dimension.
The key is in the fact that the sparsification problem for convolutional layers can be interpreted as a sparse entropic (linear) regression task.

The proposed method is tested on several benchmarks such as MNIST and CIFAR-10 and on popular CNN architectures such VGG-16 and ResNet18. For VGG-16 on CIFAR-10 one can remove $88\%$ of parameters with minimal loss in accuracy. For ResNet18 on CIFAR-10 one can remove $73\%$ of parameter with $0.5\%$ loss in performance with relatively short fine tuning.
Moreover, our method was evaluated in terms of the method for optimal network architecture search and on the example of VGG-16 it is able to discover the meaningful compressed architectures from the baseline model.

What remains the open question is finding the optimal configuration of hyperparameters to achieve the desired level of sparsity. Despite having two hyperparameters, grid search is still prohibitive for large datasets. One solution could be in Bayesian hyperparameter optimization.
Additionally, our sparsification method works well with relatively small number of data points ($T=500$ for CIFAR-10). Systematic analysis on how to select $T$ to optimize between overall approximation and the run-time belongs is needed.
Finally, robustness of baseline and sparsified models to adversarial attacks is a point worth investigating. It is known that smaller models are less prone to overfitting than the larger ones. Hence, one would expect that another benefit of the sparsification would be more robust models.

\backmatter




\section*{Declarations}

\bmhead{Ethical Approval}
Not applicable.

\bmhead{Competing interests} 
The authors have no competing interests to declare that are relevant to the content of this article.
 
\bmhead{Authors' contributions} 
Conceptualization: T.B., I.H.; Methodology: T.B.; Formal analysis and investigation:  T.B.; Writing - original draft preparation: T.B.; Writing - review and editing: I.H.; Funding acquisition: I.H.; Supervision: I.H.

\bmhead{Availability of data and materials} 
The datasets used in this dataset are open-sourced and publicly available. 

\noindent
If any of the sections are not relevant to your manuscript, please include the heading and write `Not applicable' for that section. 


\begin{appendices}

\section{Interpreting convolutional layer as linear layer}
\label{conv:layer:linear}

Let $vec$ denote (row-by-row) vectorization operation.
Then,  
$vec(Q_{i,j}) \in \R^{k^2}.$
Using vectorization and shift operator, convolution from (\ref{eq:discrete:conv}) can be interpreted as a linear layer in each point $(x_1, x_2)$ from the image domain 

$$ \big(X_{i,t} \ast Q_{i,j}\big)(x_1,x_2) = \Big\langle A(X_{i,t})(x_1,x_2) \text{ , }  vec(Q_{i,j}) \Big\rangle, $$

where $A$ returns all neighbours of $(x_1,x_2)$ with distance up $k$ in $||\cdot||_\infty$, i.e.
$$A(X_{i,t})(x_1,x_2)= \Big( X_{i,t}(x_1-k/2,x_2-k/2) , \cdots, X_{i,t}(x_1+k/2,x_2+k/2) \Big) \in \R^{k^2}.$$
Applying that to (\ref{convolutional:layer}) it follows
\begin{align*}
    Y_{j,t}(x_1,x_2) &= \sum_{i=1}^D \Big(X_{i,t} \ast Q_{i,j}\Big)(x_1,x_2) = \sum_{i=1}^D \Big\langle A(X_{i,t})(x_1,x_2) \text{ , } vec(Q_{i,j}) \Big\rangle\\
    &= \Big\langle A(X_{t})(x_1,x_2) \text{ , } Q^{vec}_{j} \Big\rangle = \big(Q^{vec}_{j}\big)^\intercal A(X_{t})(x_1,x_2),
\end{align*}
where $Q^{vec}_{j} = \Big(vec(Q_{1,j}), \cdots, vec(Q_{D,j}) \Big) \in \R^{ k^2 D}$ is a vectorized kernel for a single output channel and  
$A(X_{t})(x_1,x_2) = \Big(A(X_{1,t})(x_1,x_2), \cdots, A(X_{D,t})(x_1,x_2) \Big) \in \R^{k^2 D}$ 
is vector $k^2$ neighbours of $(x_1,x_2)$ across all $D$ input channels.
\\
This can be further written as matrix multiplication across all $M$ output channels for the single point $(x_1,x_2)$

\begin{equation}
\label{conv:layer:simplification2}
     Y_{t}(x_1,x_2) = Q^{vec} \text{ }A(X_{t})(x_1,x_2),
\end{equation}

where $Q^{vec} = \Big(Q^{vec}_{1}, \cdots, Q^{vec}_{M} \Big) \in \R^{  M\times (k^2 D)}$ and $ A(X_{t})(x_1,x_2) \in \R^{k^2 D}$. 

In the context of convolutional layers, every point from domain $H\times W$ becomes one data point to solve the system of linear equations, i.e. we can say that the actual number of data points is $T^* = (H\times W) \times T$. 
For that reason, in the next sections we will use the simplified notation which can be achieved through the variable substitution: $ (t,x_1,x_2) \to t$, $A(X_{t})(x_1,x_2) \to X_t$, and $T^* \to T$.

\section{SPARTAn algorithm}
\label{spartan_recap}
Sparse Probabilistic Approximation for Regression Task Analysis (SPARTAn) \cite{horenko2023cheap2} is a method for solving a non-linear regression problem. 
It is based on the partition of the feature space with $K$ discretization boxes and approximation with $K$ piecewise linear regression problems with entropic regularization on the feature space. 
SPARTAn requires four variables $(w,\Lambda,\gamma, C)$ which are determined by numerical minimization of the loss function $\mathcal{L}_{SPARTAn}$.
Matrix $C \in \R^{D\times K}$ represents the box centers of the feature space partition in $K$ boxes and a probability vector $\gamma_{\cdot,t} \in \R^K$ represents a probability of  $X_t \in \R^D$ belonging to any of $K$ discretization boxes, where $X_t$ is any input point from the training set.
Probability vector $w\in \R^D$ and entropy regularization enables feature selection. Finally, linear regression for every partition on $\{(X_t,Y_t)\text{ }|\text{ }t=1,\cdots,T\}$ is solved for $\Lambda \in \R^{M\times D \times K}$.

Loss function for SPARTAn is defined as
\begin{align*}
&\mathcal{L}_{SPARTAn}(w, \Lambda, C, \gamma) = \underbrace{\frac{1}{TK}\sum_{t,d,k}^{T,D,K} w_d \gamma_{k,t} (X_{d,t} - C_{d,k})^2}_{\text{Box discretization distance }} +  \epsilon_w\underbrace{\sum_{d=1}^D w_d \log w_d }_{\text{Entropy}}+ \\
&+
\epsilon_{l_2} \underbrace{\sum_{m,d,k}^{M,D,K} \Lambda^2_{d,m,k}}_{L_2} + \epsilon_r \underbrace{\frac{ 1}{TM}\sum_{t,m,k}^{T,M,K} \gamma_{k,t}\Big( Y_{m,t} - \Lambda_{0,m,k} - \sum_{d=1}^D w_d \Lambda_{m,d,k}X_{d,t} \Big)^2}_{MSE}.
\end{align*}

For given hyperparameters $(\epsilon_w, \epsilon_{l_2}, \epsilon_r, K)$, the loss function needs to be optimized in following variables
\begin{itemize}
    \item $w \in \R_{\geq 0}^D, \sum_{i=1}^D w_i = 1$ feature sparsification (probability) vector,
    \item $\Lambda \in \R^{M\times (D+1) \times K}$ regression matrix for every feature space discretization box $k\in\{1,...,K \}$,
    \item $C \in \R^{D\times K}$ centers of the feature space discretization boxes,
    \item $\gamma \in \R_{\geq 0}^{K\times T}, \sum_{k=1}^K \gamma_{k,t}= 1$ for all $t \in \{ 1, ..., T\}$: probability vector for feature $X_t$ to belong to the discretization boxes $k\in\{1,...,K \}$.
\end{itemize}

Minimization of the loss function is done via a four-step algorithm.
In each step loss function is minimized in one of the four variables, while the remaining three are kept fixed. This minimization algorithm is monotonically convergent \cite{horenko2023cheap2}. For variables $\Lambda, C,$ and $\gamma$ this step of the algorithm is analytically solvable. For $w$ a convex optimization problem needs to be solved using a matrix of second derivatives.
Optimal configuration of hyperparameters $(\epsilon_w, \epsilon_{l_2}, \epsilon_r, K)$ are determined with grid-search on discretized hyperparameter space on the validation set.

\subsection{Sparse entropic regression using SPARTAn algorithm}
\label{sec:fcc:layer:sparse}


Linear adaptation of SPARTAn is defined by setting $K=1$, as we aim to solve a sparse linear regression problem motivated by (\ref{conv:layer:simplification}). Hence, parameters related to feature space partition $\gamma$ and $C$ are not required and our sparsification loss becomes
\begin{align}
\label{sparse_loss}
&\mathcal{L}_{sparsify}(w, \Lambda) =  \epsilon_w\underbrace{\sum_{d=1}^D w_d \log w_d}_{\text{Entropy}} + 
\epsilon_{l_2} \underbrace{\sum_{m,d=1}^{M,D+1} \Lambda^2_{m,d-1}}_{L_2} 
+ \underbrace{\frac{1}{TM} \sum_{t,m=1}^{T,M} \Big( Y_{m,t} - \Lambda_{m,0} - \sum_{d=1}^D w_d \Lambda_{m,d}X_{d,t} \Big)^2}_{MSE}.
\end{align}
Loss function $\mathcal{L}_{sparsify}$ consists of three terms: mean square or approximation error or (MSE), $L_2$ regularization on regression weights $\Lambda$, and entropic regularization on probability vector $w$.
The minimization of $\mathcal{L}_{sparsify}$ solves for regression matrix $\Lambda \in  \mathds{R}^{M\times (D+1)}$ and probability vector $w=(w_1,...,w_D) \in \R^D_{\geq 0}$, $\sum_{i=1}^D w_i = 1$ with respect to chosen hyperparameters $(\epsilon_w, \epsilon_{l_2})$. 
Vector $w$ has a probabilistic interpretation as a feature selection (or importance) vector on $X_t$.
Feature dimensions, whose vector components $w_i$ are close to zero (e.g. $<10^{-6}$), have no effect on the approximation error (MSE) and hence can be discarded as less relevant. The sparsity of the vector $w$ is ensured from the minimum entropy principle in entropy regularization, i.e. $\epsilon_w < 0$. 


\section{Optimization steps for $\mathcal{L}_{sparsify}$}
\label{optimize_sparsification}

Optimization problem then becomes
\begin{align*}
\mathcal{L}_{sparsify}(w, \Lambda) &=  \epsilon_w\underbrace{\sum_{d=1}^D w_d \log w_d}_{\text{Entropy}} + 
\epsilon_{l_2} \underbrace{\sum_{m,d=1}^{M,(k^2D+1)} \Lambda^2_{m,d-1}}_{L_2}+ \\
&+ \underbrace{\frac{1}{TM} \sum_{t,m=1}^{T,M} \Bigg( Y_{m,t} - \Lambda_{m,0} - \sum_{d=1}^D w_d \bigg(\sum_{l=1}^{k^2} \Lambda_{m,(d-1)k^2+l}X_{(d-1)k^2+l,t} \bigg) \Bigg)^2}_{MSE}.
\end{align*}

In the first step probability vector $w \in \R^D$ is initialized uniformly, i.e. $w=(1/D,...,1/D)$.

\subsection{$\Lambda$-step}
In this step, $w$ is kept fixed and the loss is minimized in $\gamma$. As entropy regularization is only applied on $w$, it acts as a constant in this step and is hence omitted

\begin{align*}
\mathcal{L}_{sparsify}(\Lambda) &=  
\epsilon_{l_2} \sum_{m,d=1}^{M,(k^2D+1)} \Lambda^2_{m,d-1}
+\frac{1}{TM} \sum_{t,m=1}^{T,M} \Bigg( Y_{m,t} - \Lambda_{m,0} - \sum_{d=1}^D w_d \bigg(\sum_{l=1}^{k^2} \Lambda_{m,(d-1)k^2+l}X_{(d-1)k^2+l,t} \bigg) \Bigg)^2 =
\\
&= \frac{1}{M}\sum_{m=1}^M \Big( \epsilon_{l_2}  \sum_{d=1}^{(k^2D+1)} \Lambda^2_{m,d-1}
+ \frac{1}{T} \sum_{t}^{T} \Big( Y_{m,t} - \Lambda_{m,0} - \sum_{d=1}^D \sum_{l=1}^{k^2} \Lambda_{m,(d-1)k^2+l} \underbrace{w_d X_{(d-1)k^2+l,t}}_{Z_{(d-1)k^2+l,t}}  \Big)^2 \Big)\\
\end{align*}
for $Z=(Z_{i,j}) \in \R^{(k^2 D) \times T}$ where $Z_{(d-1)k^2+l,t} = w_d X_{(d-1)k^2+l,t}$. 
\\
Hence, this can be interpreted as $M$ ridge regression problems for every row of the matrix $\Lambda$. This is analytically solvable, i.e. let $Z_1=(1_T,Z^\intercal) \in \R^{T\times(k^2D+1) }$, then  
$$\Lambda_{m, \cdot} = (Z_1^\intercal Z_1+ \epsilon_{l_2}I_{k^2D+1})^{-1} Z_1^\intercal Y^{(m,\cdot)}$$
for $m=1,...,M$.

\subsection{$W$-step}
In this step $\Lambda$ is kept fixed, and the loss is optimized in $w$. As $L_2$ regularization is dependent only on $\Lambda$, it acts as a constant in this step and hence can be omitted: 

\begin{align*}
    \mathcal{L}_{sparsify}(w) &=  \epsilon_w\sum_{d=1}^D w_d \log w_d + \frac{1}{TM} \sum_{t,m=1}^{T,M} \Bigg( Y_{m,t} - \Lambda_{m,0} - \sum_{d=1}^D w_d \bigg(\sum_{l=1}^{k^2} \Lambda_{m,(d-1)k^2+l}X_{(d-1)k^2+l,t} \bigg) \Bigg)^2.
\end{align*} 

Simplifying this 
\begin{align*}
    \mathcal{L}_{sparsify}(w) & =  \epsilon_w \langle w, \log w \rangle  +
 \frac{1}{TM} \sum_{t,m=1}^{T,M}  \Bigg(\sum_{d=1}^D  w_d \bigg(\sum_{l=1}^{k^2} \Lambda_{m,(d-1)k^2+l}X_{(d-1)k^2+l,t} \bigg) \Bigg)^2 +\\
 &- \frac{2}{TM} \sum_{t,m=1}^{T,M} \Big( Y_{m,t} - \Lambda_{m,0}\Big)\Big( \sum_{d=1}^D w_d \big(\sum_{l=1}^{k^2} \Lambda_{m,(d-1)k^2+l}X_{(d-1)k^2+l,t} \big) \Big) + \underbrace{\frac{1}{TM}\sum_{t,m=1}^{T,M}(Y_{m,t}-\Lambda_{m,0})^2}_{\text{independent of } w \to omit} = \\
 &=    
 \frac{1 }{TM}  \sum_{d_1=1}^D \sum_{d_2=1}^D \sum_{m=1}^M w_{d_1} \Big\langle  \sum_{l_1}^{k^2} \Lambda_{m,(d_1-1)k^2+l_1}X_{(d_1-1)k^2+l_1,.} ,    \sum_{l_2}^{k^2}  \Lambda_{m,(d_2-1)k^2+l_2}X_{(d_2-1)k^2+l_2,.} \Big\rangle_T w_{d_2}+ \\
 & -\frac{2}{TM} \sum_{m,d=1}^{M,D} w_d \Big\langle  Y_{m,.} - \Lambda_{m,0} , \sum_{l=1}^{k^2}\Lambda_{m,(d-1)k^2+l}X_{(d-1)k^2+l,.} \Big\rangle_T + \epsilon_w \langle w, \log w \rangle .
\end{align*}

This is solved using MathWorks' \textit{fmincon} with constraints $\langle w, (1,...,1)^\intercal \rangle = 1$ and $0 \leq I_D w \leq 1$.

\section{Details on networks}
\label{ref:net:details}
Details on network are given for LeNet (Table \ref{tab:lenet}), VGG-16 (Table \ref{tab:VGG-overview}), and ResNet18 (Table \ref{tab:resnet18-overview}).

\begin{table}[]
    \centering
    \begin{tabular}{|c|c|c|c|c|}
         \hline
         Layer & \#input channels & \#output channels & output size & \#parameters (\% of total) \\
         \hline
         conv0 & 1 & 6 & $28\times 28$ & 156 (0.24\%)\\
         avg pool & 6 & 6 & $14\times 14$ & - \\
         \hline
         conv1 & 6 & 16 & $10\times 10$ & 2,416 (3.91\%)\\
         avg pool & 16 & 16 & $5 \times 5$ & -\\
         flatten & 16 & 400 & $1 \times 1$ & - \\
         \hline
         fc1 & 400 & 120 & $1\times 1$ & 48,120 (77.98\%)\\
         fc2 & 120 & 84 & $1\times 1$ & 10,164 (16.47\%) \\
         fc3 & 84 & 10 & $1\times 1$ & 850 (1.37\%) \\
         \hline
    \end{tabular}
    \caption{LeNet: architecture overview. Total number of parameters $61,706$. }
    \label{tab:lenet}
\end{table}

\begin{table}[]
    \centering
\begin{tabular}{|c|c|c|c|c|}
     \hline
     Layer & \#input channels & \#output channels & output size & \#parameters (\% of total) \\
     \hline
     conv0 & 3 & 64 & $32 \times 32$ & 1,792 (0.01\%) \\ 
     conv1 & 64 & 64 & $32 \times 32$ & 36,928 (0.25\%) \\  
     max pool & 64 & 64 &  $16 \times 16$ & -  \\ 
     \hline 
     conv2 & 64 & 128 & $16 \times 16$ & 73,856 (0.5\%) \\ 
     conv3 & 128 & 128 & $16 \times 16$ &  147,584 (1\%) \\ 
     max pool & 128 & 128 & $8 \times 8$ & - \\
     \hline 
     conv4 & 128 & 256 & $8 \times 8$  &  295,168 (2\%)\\ 
     conv5 & 256 & 256 & $8 \times 8$ &  590,080 (4\%)\\ 
     conv6 & 256 & 256 & $8 \times 8$  &  590,080 (4\%) \\  
     max pool & 256 & 256 & $4 \times 4$ & - \\
     \hline 
     conv7 & 256 & 512 & $4 \times 4$ & 1,180,160 (8.01\%) \\ 
     conv8 & 512 & 512 & $4 \times 4$ & 2,359,808  (16.02\%)\\ 
     conv9 & 512 & 512 & $4 \times 4$ & 2,359,808 (16.02\%)\\ 
     max pool & 512 & 512 & $2 \times 2$  & - \\
     \hline 
     conv10 & 512 & 512 & $2 \times 2$ &  2,359,808 (16.02\%)\\ 
     conv11 & 512 & 512 & $2 \times 2$ & 2,359,808 (16.02\%)\\ 
     conv12 & 512 & 512 & $2 \times 2$ &  2,359,808 (16.02\%)\\ 
     max pool & 512 & 512 & $1 \times 1$ & - \\
     \hline
     fc & 512 & 10 & $1 \times 1$ &  5,130 (0.03\%) \\ 
     \hline 
\end{tabular}
    \caption{Overview of the architecture of VGG-16 trained on CIFAR-10 dataset. Compactly, this can be written using only number of outputs channels: $$64, 64, \text{M}, 128, 128, \text{M}, 256, 256, 256, \text{M}, 512, 512, 512, \text{M}, 512, 512, 512, \text{M}$$ where 'M' denotes max pooling and where fully connected layer has been omitted. Convolutional layers have in total $14,719,818$ parameters, as they defined on $3\times 3$ window. Additional $8,448$ parameters arise from affine batch normalization. Hence, the total number of parameters is $14,728,266$.}
    \label{tab:VGG-overview}
\end{table}

\begin{table}[]
    \centering
\begin{tabular}{|c|c|c|c|c|}
     \hline
     Layer & \#input channels & \#output channels & output size & \#parameters (\% of total) \\
     \hline
     conv0 & 3 & 64 & $32 \times 32$ &  1,728 (0.02\%) \\ 
     \hline
     conv1 & 64 & 64 & $32 \times 32$ &  36,864 (0.33\%) \\  
     conv2 & 64 & 64 &  $32 \times 32$ &  36,864 (0.33\%) \\
     conv3 & 64 & 64 &  $32 \times 32$ &  36,864 (0.33\%) \\ 
     conv4 & 64 & 64 &  $32 \times 32$  &  36,864 (0.33\%) \\ 
     \hline
     conv5 & 64 & 128 & $16 \times 16$  & 73,728 (0.65\%) \\  
     conv6 & 128 & 128 & $16 \times 16$ &  147,456 (1.3\%) \\ 
     conv5 shortcut & 64 & 128 & $16 \times 16$ &  8,192 (0.07\%) \\
     conv7 & 128 & 128 & $16 \times 16$ &  147,456 (1.3\%) \\ 
     conv8 & 128 & 128 & $16 \times 16$ &  147,456 (1.3\%) \\ 
     \hline 
     conv9 & 128 & 256 & $8 \times 8$  & 294,912 (2.6\%) \\  
     conv10 & 256 & 256 & $8 \times 8$ &  589,824 (5.2\%) \\ 
     conv9 shortcut & 128 & 256 & $8 \times 8$ &  32,768 (0.03\%) \\  
     conv11 & 256 & 256 & $8 \times 8$ &  589,824 5.2\%) \\ 
     conv12 & 256 & 256 & $8 \times 8$ &  589,824 (5.2\%) \\ 
     \hline  
     conv13 & 256 & 512 & $4 \times 4$  &  1,179,648 (10.5\%) \\  
     conv14 & 512 & 512 & $4 \times 4$ & 2,359,296 (21.1\%) \\ 
     conv13 shortcut & 256 & 512 & $4\times4$ & 131,072(1.2\%) \\ 
     conv15 & 512 & 512 & $4 \times 4$ &  2,359,296 (21.1\%) \\ 
     conv16 & 512 & 512 & $4 \times 4$ &  2,359,296 (21.1\%) \\
     \hline 
     avg pool & 512 & 512 & $1\times 1$ & - \\
     fc & 512 & 10 & $1\times 1$ & 5,140 (0.046\%) \\
     \hline
\end{tabular}
    \caption{ResNet18 has in total 18 convolutional layers (+3 shorcut convolutional layers) and in total $11,173,962$ parameters. Convolutions with shortcut represent $1\times 1$ (in contrast to the remaining ones which are $3\times 3$). All of the convolutions do not use bias, however the batch normalizations, which succeed convolutional layers, are affine and hence, introduce additional $9,600$ parameters.}
    \label{tab:resnet18-overview}
\end{table}

\begin{figure}
    \centering
    \includegraphics[width=0.55\textwidth]{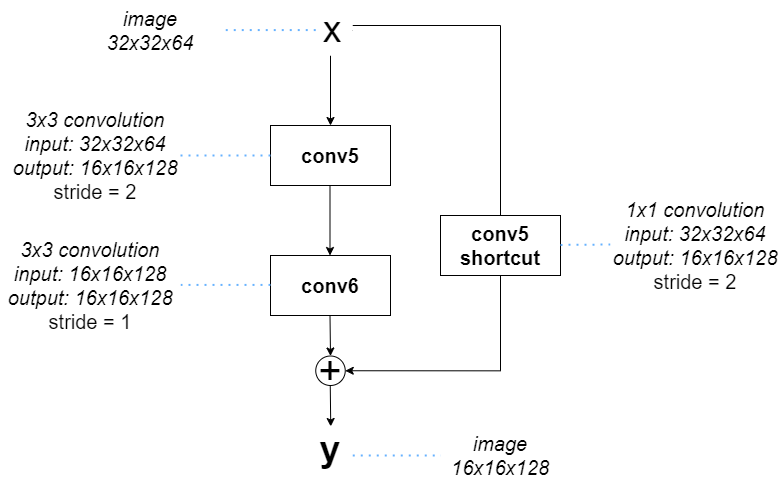}
    \caption{Illustration of one convolutional block with the residual connection from ResNet18 (Table \ref{tab:resnet18-overview}): conv5, conv6, and conv5 shortcut. Here, every "conv" box includes convolution, batch normalization, and non-linearity ReLU. Note that "conv5 shortcut" is needed to the match the number of chanells (64) of the output of "conv6", as $x$ has 32 channels. In cases when the number of channels match one avoids using this operator in the residual connection.}
    \label{fig:resblock}
\end{figure}




\end{appendices}


\bibliographystyle{elsarticle-num}
\bibliography{lib}

\end{document}